# Provenance-grouped protocol for evaluating machine learning in perovskite photovoltaics

Qiwei Han[*1,2], Chi Zhou[2]
[1]Department of Chemistry, Duke University, Durham, NC 27708, USA
[2]Department of Computer Science, Georgia Institute of Technology, Atlanta, GA 30332, USA
[*]Corresponding author.

## Abstract

Machine-learning models for perovskite solar cells are typically evaluated using random data splits, which we show systematically leak information through massive internal duplication. Because nearly two-thirds of curated devices share their exact recipe with another, random splits primarily reward the mere recall of historical data rather than testing true physical generalisation. To solve this, we mapped the institutional provenance of the Perovskite Database and established a rigorous hierarchy of grouped splits that progressively isolate duplicate devices, publications, and institutions. Benchmarking thirteen models establishes new state-of-the-art performance across all splits and reveals a critical regime shift: while tree ensembles dominate random splits by recognising familiar recipes, their ranking collapses on genuinely unseen institutions and future discoveries, where composition-aware neural architectures excel. Furthermore, feature importance demonstrates that the device architecture and fabrication methodology dictate final efficiency just as strongly as the absorber chemistry. This standardised protocol empowers the community to separate memorisation from true physical discovery.

## Introduction

Machine learning is increasingly used to predict experimental outcomes, identify patterns in complex data, and guide experiments across scientific disciplines [1–8], and large archives of experimental records have been assembled to support this effort. Perovskite solar cells provide a striking example. Their efficiency rose from 3.8% to above 26% within a decade [9–12], and the rapid pace of the field produced the Perovskite Database [13]. With more than 42,000 devices from thousands of publications, it is the largest device archive in photovoltaics.

The rapid development of the field also shapes the contents of this archive. Many publications report several cells manufactured in the same way, and later studies often replicate the designs of earlier ones. Consequently, many records share exactly the same recorded inputs. When such an archive is split at random, identical devices end up in both the training and test sets. A model can then accurately predict a test device's efficiency simply by recalling its duplicate in the training set, rather than by learning the underlying relationship between device properties and efficiency. This form of data leakage has been documented across seventeen fields and 294 papers [14].

Numerous models have been trained on the Perovskite Database, with their reported accuracies improving steadily [15–19]. However, each model was evaluated on its own selection of records, mainly with random splits, and typically compared only against other neural models [16, 19]. As a result, the reported accuracies cannot be directly compared, each includes an unknown contribution from recalled duplicates, and the performance of simple baselines on the same data remains unknown. The standard remedy for this issue is a grouped split, which ensures that all records sharing a specific trait, such as originating from the same publication or institution, are kept on the same side of the split [20, 21]. Implementing a grouped split requires knowing the publication and laboratory of origin for each record. While the archive includes the DOI for each publication, it lacks explicit laboratory identifiers.

In this work, we evaluate models of the Perovskite Database on genuinely unseen devices. We construct a hierarchy of grouped splits that progressively separates duplicate devices, publications, and institutions, allowing the leakage at each level to be measured independently; a temporal split by publication date complements this hierarchy. To achieve the strictest level of evaluation, we resolved the provenance of the archive's records, identifying the corresponding author's institution for 94.7% of its 4,683 publications (796 institutions). This extends cross-laboratory evaluation, previously limited to just 819 cells from three laboratories [22], to the full, large-scale public archive, and turns the bibliographic confounding shown across materials tasks [23] into a rigorous protocol that removes it.

Using this protocol, we first quantify the extent of duplication within the archive and show how strongly a random split rewards the recall of duplicates. When the evaluation setup of the published state of the art [16] is replicated, a lookup table that does no learning, predicting each device from its duplicates in the training set, simply reaches a mean absolute error (MAE) of 2.934%, below the reported state-of-the-art value of 2.936% [16]. Accuracy on a random split therefore reflects the recall of duplicates as much as the model itself, and the protocol is designed to separate recall from generalisation to new devices. We then establish the first common benchmark of thirteen models, trained and tested on identical partitions of the curated archive, and set a new state of the art under every split, with MAEs of 1.973% on a random split, 2.396% once duplicates are kept apart, 2.832% on new publications, 2.912% on new institutions and 3.192% on publications from later years. Over ten further partitions of each split, we build the rank correlation of the models between a random partition and a partition of each split, and show that a model's rank on a random split carries no information about its rank on new institutions. Finally, feature-importance analysis quantitatively identifies the absorber composition, the device architecture and the fabrication process as the inputs that carry the predictive information, and reveals the limitation of tree-based models in interpreting materials chemistry and extrapolating to new devices, which points to the future direction of model optimisation. Together, this hierarchy of splits, provenance mappings and benchmark forms a standardised protocol against which the community can compare models on equal terms, separate recall from generalisation, and direct model development towards the inputs that carry physical information.

## Data Curation and Split Protocols

We first curated the Perovskite Database, leaving 23,620 physical devices from the original 42,443 records. We selected features that dictate the physical performance of perovskite solar cells: the absorber composition, the sequence of transport layers, the fabrication procedure (including solvents and annealing), the measurement conditions, and the measured performance results. Table S1 provides the missing-value analysis for these core fields; we kept only records that are completely populated across them. To capture these recipes comprehensively, we expanded the raw archive records into 24 descriptive fields (Table S2). The full curation and expansion details are described in data curation of Methods.

To evaluate models on genuinely unseen devices, we established a hierarchy of grouped splits based on three types of recurring relationships: an identical description (duplicates), a shared publication, and a shared institution (Fig. 1). Publication and institution are used strictly to form the splits and are never passed as inputs to the models.

The protocol defines the following splits:

**Random split.**

Devices are assigned independently, as in earlier work on this archive.

**Level 1, feature-grouped.**

All duplicates of a device are assigned to the same set.

**Level 2, publication-grouped.**

All devices from a single publication are assigned to the same set.

**Level 3, institution-grouped.**

All devices from a single institution are assigned to the same set, ensuring no test device shares its institution with a training device.

**Temporal split.**

Publications up to 2016 form the training set, 2017 the validation set, and 2018 onward the test set.

Detailed statistics on how these splits apply to the curated archive are provided in the split definitions of Methods. The results and their meaning will be discussed in later sections.

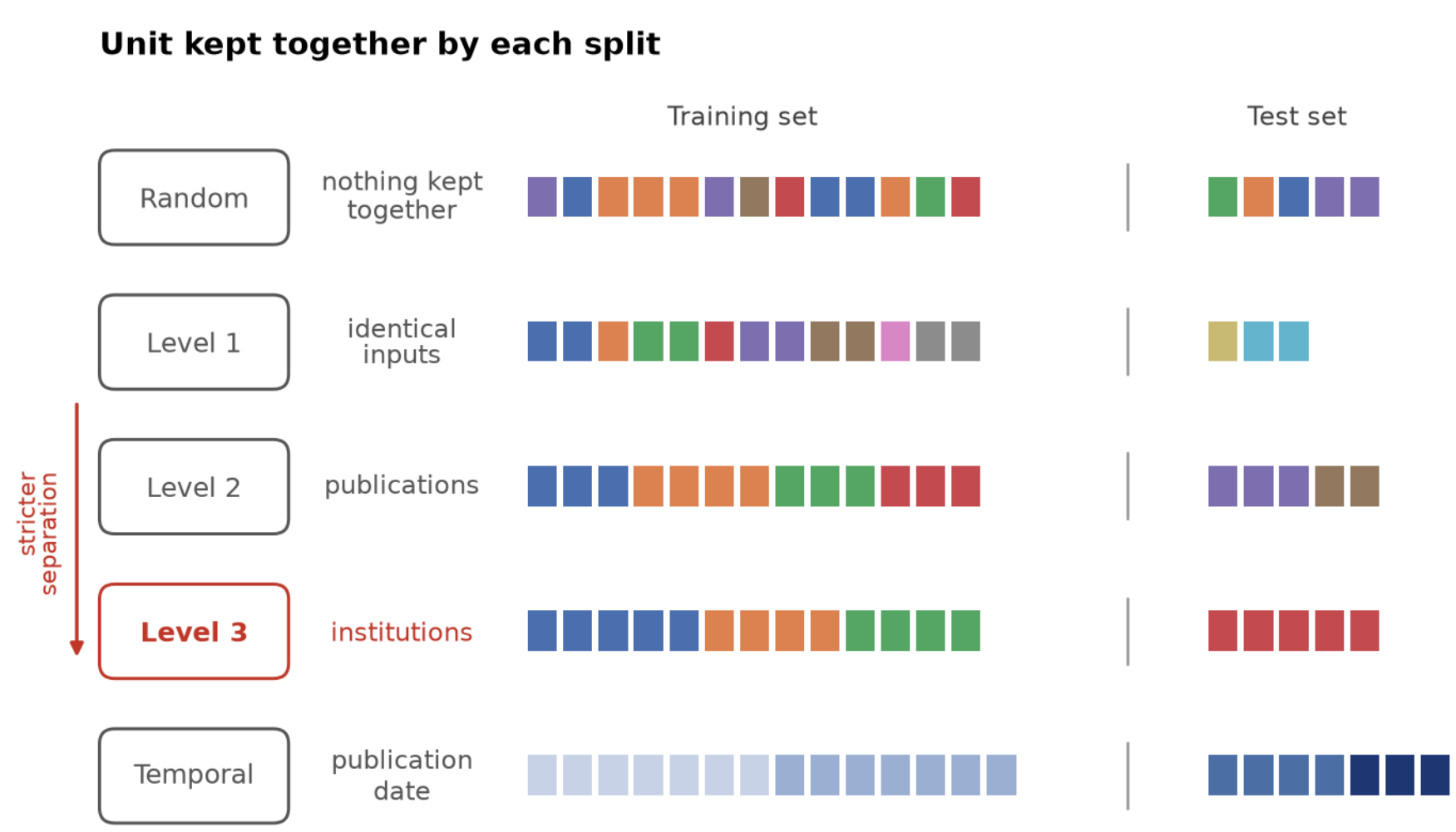

**The provenance-grouped protocol.** The splits and the group that each keeps on one side. Each square is a device, coloured by the group that the split keeps together. In the random split the same colour appears in both the training and the test set; at levels 1 to 3 no colour does, and the groups grow from duplicates to publications and then to institutions. The temporal split assigns publications by date (shade). Level 3 is the only split in which no institution contributes to both sets. The publication and institution of each device are recovered as described in provenance mapping of Methods.

## Results and discussion

### Progressive elimination of data leakage

The necessity of the grouped protocol becomes apparent when examining the archive's massive internal duplication. Most devices in the curated archive have a duplicate. The 23,620 curated devices carry only 12,933 distinct descriptions, and 62.3% of them share their description with at least one other device (Fig. 2a). Duplicates are separately made cells whose efficiencies differ (the PCE of the 60 cells in the largest group ranges from 13.48% to 16.53%; details in curation of Methods, Fig. S1), but they almost always come from one publication: only 20 of the 4,024 groups of duplicates span more than one, and 2,744 of them (68.2%, holding 61.4% of their devices) are identical in all 249 archive columns that describe how a device was made (details in duplication analysis of Methods). A group of duplicates is therefore, in most cases, a series of cells made to one recipe and reported together.

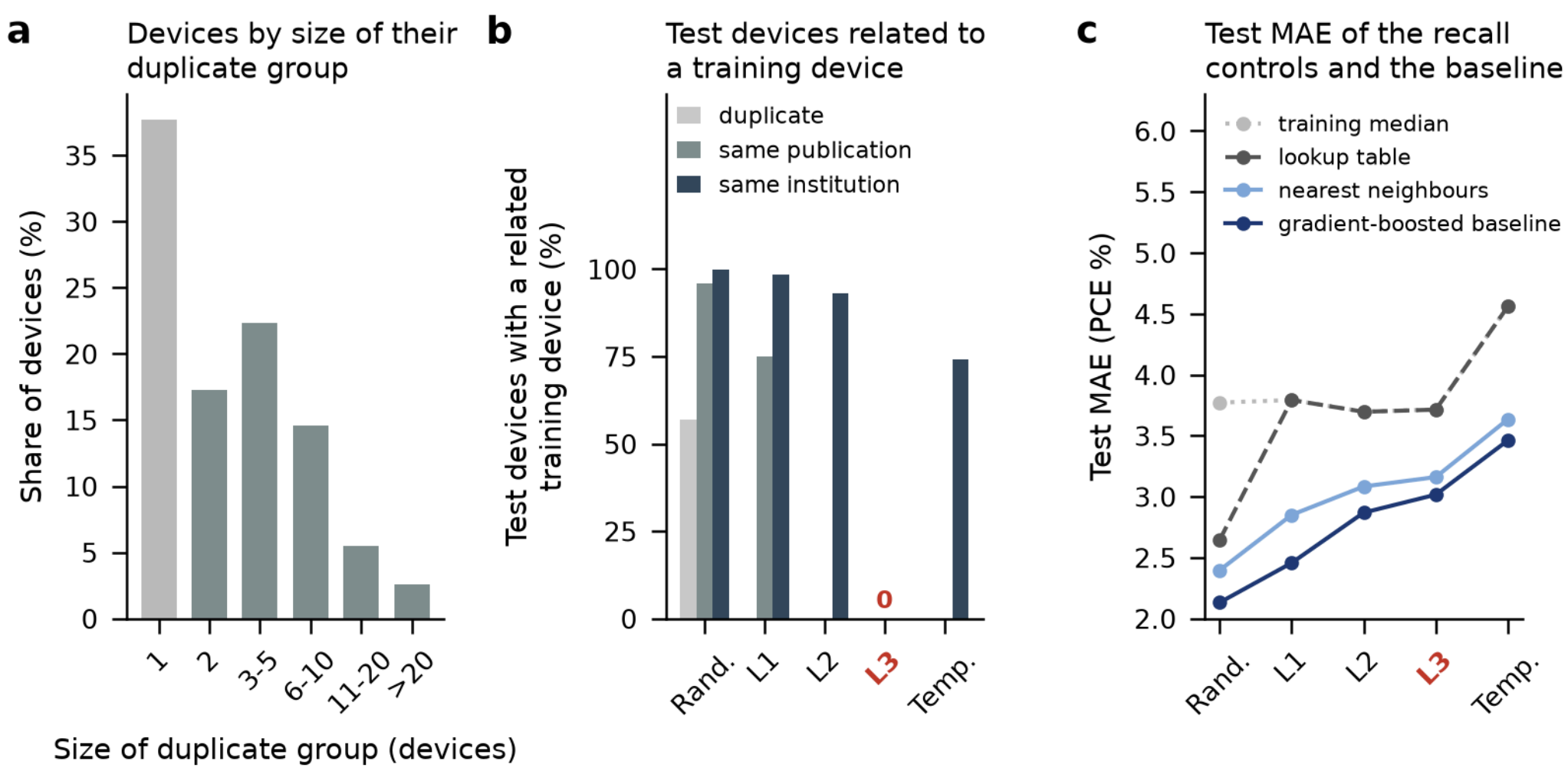

**Duplication in the archive and relatedness of test to training devices. a**, Share of the 23,620 devices by the size of their duplicate group (devices with the same description). **b**, For each split, the percentage of test devices with a duplicate in the training set (light), from a publication with devices in the training set (middle) and from an institution with devices in the training set (dark). **c**, Test MAE on every split of the lookup table, the nearest-neighbour model, the training median and the gradient-boosted baseline. Data are in Table 1 and Tables S3 and S5.

To measure how much a random split rewards the recall of duplicates, we scored a lookup table that predicts each test device as the median PCE of its duplicates in the training set, and as the training median when it has none (details in recall controls of Methods). When the archive is split at random, 56.9% of the test devices have a duplicate in the training set (Table S3), and the lookup table reaches a mean absolute error (MAE) of 2.643% (Table 1) without learning how efficiency depends on any input. To compare it on equal terms with the published state of the art [16], we replicated that evaluation setup (details in benchmark settings of Methods, Fig. S2): the lookup table reaches an MAE of 2.934%, against the reported state-of-the-art value of 2.936% [16], and gradient-boosted tree ensembles with default settings reach 2.893% to 2.918% (2.872% to 2.888% with settings selected on the validation set; Table S4). Because a random split of this archive allows such extensive recall, a model that learns nothing beyond the duplicates matches the published state of the art on it, and untuned tree ensembles, which the published comparisons did not include, surpass it. Accuracy on a random split therefore cannot show that a new architecture generalises to new devices.

Having shown that a random split rewards recall, we measured how much of this recall each grouped split removes. Each level removes one more route by which a test device can be predicted from related training devices (Table S3; Fig. 2b; details in split definitions of Methods). The duplicate and publication groups can be read directly from the archive: devices are duplicates if they share all 24 descriptive fields, and they share a publication if they have the same DOI. Because laboratory identifiers are not recorded, we mapped the corresponding-author affiliations for each publication using OpenAlex, identifying the institution for 95.8% of the curated devices (details in provenance mapping of Methods). On the random split, 95.8% of the test devices share their publication and 99.8% their institution with a training device, in addition to the 56.9% that have a duplicate there. Level 1 removes duplicates, yet 74.9% of its test devices still share a publication with the training set (883 of its 1,072 test publications appear on both sides); level 2 separates publications, yet 93.0% of its test devices still share an institution; and level 3 separates institutions, so that no test device shares a duplicate, a publication or an institution with the training set. The temporal split separates publications by date, and 74.1% of its test devices share an institution with the training set.

To measure how much accuracy recall still provides at each level, we compared the lookup table with a second recall control, a nearest-neighbour model that predicts a test device as the mean PCE of its $k$ most similar training devices, with $k$ chosen on validation (Table 1 and Table S5; Fig. 2c; details in recall controls of Methods). Because level 1 removes all duplicates, the error of the lookup table rises immediately from 2.643% on the random split to 3.793%, where it equals the error of predicting the training median. The nearest-neighbour model, however, still finds close relatives on the random split and at level 1, averaging only $k = 3$ and $k = 20$ devices.

But at levels 2 and 3, where devices from the same publication do not remain in training, it is forced to average $k$ = 100 ($k$ = 50 on the temporal split). Close relatives in the training set therefore supply the accuracy of recall, and the grouped splits successfully remove them.

Even when exact duplicates are removed, related devices from the same publication or institution still leak information. To isolate this effect, we evaluated the gradient-boosting reference model on training sets that differ only in whether they include these non-duplicate relatives. We strictly fixed the test devices, the training-set size, and ensured no duplicates crossed the split (details in leakage analysis of Methods). If different devices from the test device's publication are allowed in the training set, the mean absolute error artificially drops by 0.215% ± 0.008% (mean over twenty repetitions, each above 0.15%; 0.601% if duplicates are also allowed). If devices from the same institution are allowed, the error drops by 0.106% ± 0.008% (positive in all twenty repetitions, 0.041% to 0.170%). This performance gap is driven entirely by leakage, as merely shrinking the training set to match the stricter splits raises the error by a negligible 0.003% ± 0.004% to 0.011% ± 0.006%. Even though publication and institution identifiers are never given to the model, sharing a source provides an unfair predictive advantage because the model implicitly learns the hidden systematic biases of that specific institution. These hidden advantages confirm why random splits are insufficient for physical datasets: grouped splits are rigorously essential to evaluate whether a model has actually learned generalizable physics. Level 2 of our protocol removes the publication advantage, and level 3 removes both.

**The thirteen-model benchmark**

Having established level 3 as the strictest test and tree ensembles as strong baselines, we benchmarked thirteen models on every split to set the state of the art under each kind of generalisation: six tree-based models, six neural models and a gradient-boosted baseline with fixed settings (Table 1; Fig. 3a,b; details in models of Methods). Two of the neural models read the recorded names of the device layers with a frozen scientific language model, SciBERT [24]. To guarantee a strictly fair comparison, no test device was ever used to tune hyperparameters, to stop training, or to select models (details in benchmark settings of Methods). By evaluating all thirteen models strictly on identical data, this benchmark establishes the first comprehensive state-of-the-art baselines for perovskite solar cells under rigorous level-2, level-3 and temporal evaluation.

**Test MAE (PCE, %) of the thirteen models and the two recall controls on every split.** The splits are defined in Section 2. The recall controls predict a test device from its duplicates in the training set (lookup table) or from its most similar training devices (nearest neighbours). Bold marks the lowest error on each split. Models are ordered by error at level 3.

| Model | Type | Random | Level 1 | Level 2 | Level 3 | Temporal |
|---|---|---|---|---|---|---|
| SciBERT+ attention | neural | 1.993 | 2.454 | 2.871 | **2.912** | 3.228 |
| CatBoost (native) | tree | 2.171 | 2.536 | 2.839 | 2.933 | 3.440 |
| XGBoost | tree | 2.027 | **2.396** | **2.832** | 2.949 | 3.471 |
| LightGBM | tree | **1.973** | 2.409 | 2.848 | 2.957 | 3.476 |
| SciBERT+ MLP | neural | 2.080 | 2.563 | 2.901 | 2.965 | 3.259 |
| FT-Transformer | neural | 2.078 | 2.701 | 2.962 | 2.977 | 3.215 |
| TabM | neural | 2.041 | 2.545 | 2.965 | 2.978 | 3.246 |
| MLP | neural | 2.065 | 2.683 | 2.938 | 2.992 | **3.192** |
| Histogram gradient boosting | tree | 2.025 | 2.429 | 2.863 | 3.000 | 3.521 |
| Gradient-boosted baseline | baseline | 2.131 | 2.457 | 2.869 | 3.017 | 3.462 |
| CatBoost (encoded) | tree | 2.127 | 2.480 | 2.892 | 3.018 | 3.388 |
| Random forest | tree | 2.044 | 2.481 | 2.923 | 3.022 | 3.621 |
| ResNet-MLP | neural | 2.084 | 2.718 | 3.033 | 3.032 | 3.268 |
| Lookup table | recall control | 2.643 | 3.793 | 3.696 | 3.714 | 4.560 |
| Nearest neighbours | recall control | 2.396 | 2.850 | 3.084 | 3.161 | 3.635 |

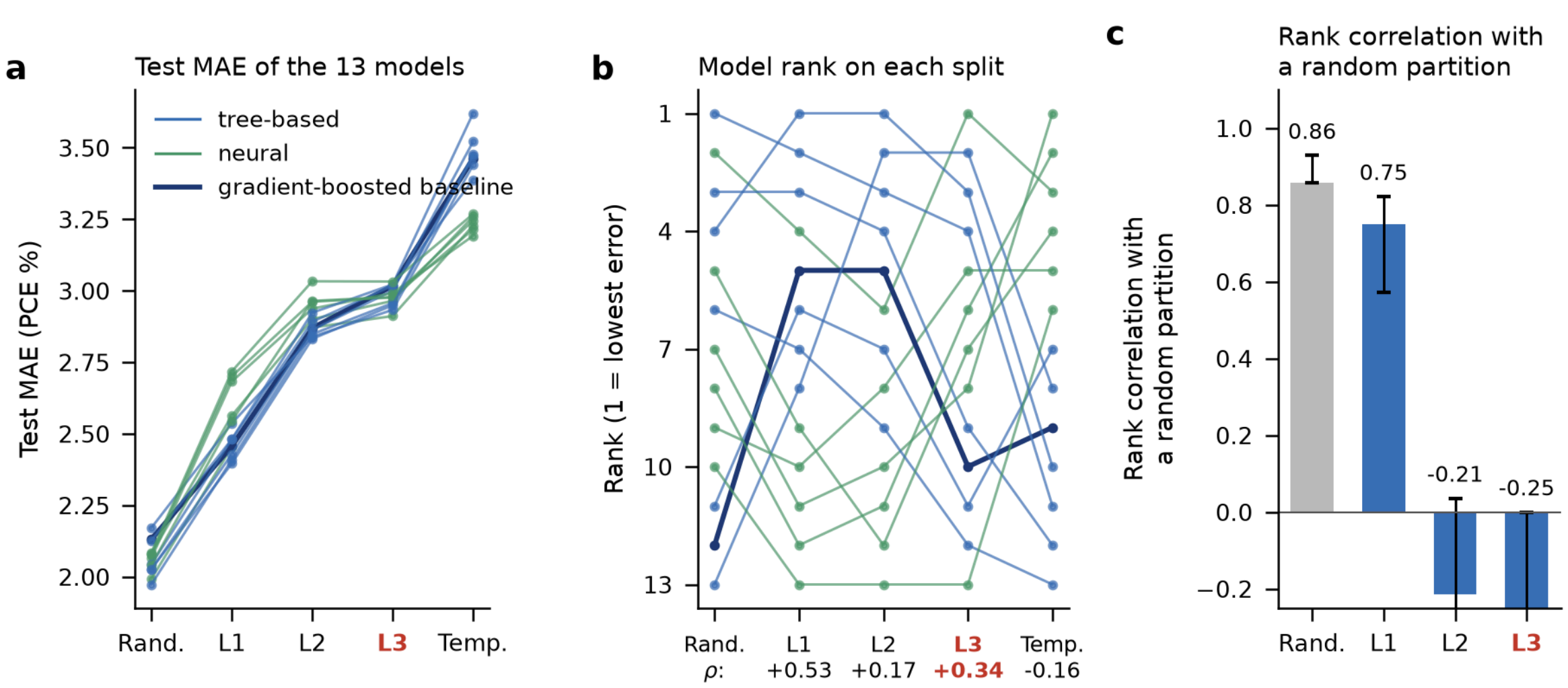

**The benchmark on the curated archive. a**, Test MAE of the thirteen models on every split, one line per model (tree-based in blue, neural in green, the gradient-boosted baseline in dark blue). **b**, Rank of each model on every split, with the Spearman correlation ρ between its rank on the random split and its rank on each other split. **c**, Spearman correlation between the ranking of the six tree-based models and the baseline on a random partition and their ranking on another random partition or on a partition of level 1, 2 or 3, over the eleven partitions of each split, shown as the median (bar and value) and interquartile range (error bar), over 55 pairs of random partitions and 121 pairs otherwise. Data are in Table 1 and Fig. S3.

To test whether a ranking measured on a random split predicts the ranking on the other splits, we computed the Spearman correlation between each model's rank on the random split and its rank on each other split (Fig. 3b). The error of every model rises from the random split to level 1 and to level 2, stays close to that level at level 3 and is highest on the temporal split (Table 1), and the order of the models changes with it: across the thirteen models, the correlation is +0.527 at level 1, +0.170 at level 2, +0.335 at level 3 and −0.159 on the temporal split. To illustrate how strongly these rankings shift: CatBoost with native categorical handling ranks last on the random split but second at levels 2 and 3, while the multi-layer perceptron (MLP) ranks seventh on the random split yet wins the temporal split. Because a random split primarily rewards the recall of duplicates, these rankings reveal that models which excel at memorization are fundamentally different from those that excel at physical extrapolation. Consequently, crowning a model as the state-of-the-art based on a random split actively misleads the field into deploying architectures that fail to generalize to new discoveries.

To test whether this loss of agreement holds beyond a single draw of each split, we built ten further partitions of the random split and of each level with the same rule, which leak in the same pattern as the published ones (Table S6), and refitted the six tree-based models and the baseline on each (details in benchmark settings of Methods, Fig. S3). On every draw, the error of every model rises from the random split to level 1 and again to levels 2 and 3. Two random partitions rank the models alike (median Spearman correlation 0.86), whereas a random partition and a partition of level 1 agree somewhat less (median 0.75), and a random partition and a partition of level 2 or 3 do not agree at all (median −0.21 and −0.25; Fig. 3c). Across the partitions, a model's error varies with an average SD of 0.05% to 0.08% at levels 1 to 3, whereas its difference from the baseline on the same partition has an average SD of only 0.02% (Fig. S3b). These tight relative differences confirm that the collapse of model rankings is not a statistical fluke, but a fundamental failure of the random split itself. Furthermore, the high volatility of absolute errors across different test draws proves that comparing absolute errors from different datasets is invalid; new architectures must be benchmarked strictly against baselines on identical data partitions to prove a true advance.

To identify which architectures are best suited for physical discovery, we compared the top-performing models across the splits (Table 1; details in models of Methods). A clear regime shift emerges. On the looser splits (random, level 1 and level 2) where training data leaks into the test set through shared recipes or publications, tree ensembles dominate. LightGBM and XGBoost win these splits by treating device layer names as opaque categorical strings, which is highly efficient for recognizing familiar recipes. However, at level 3, where every test institution is entirely unseen, the tree ensembles lose their lead. The lowest error on this strict extrapolation test (2.912%) belongs to SciBERT+attention, a neural model that uses a scientific language model to read the semantic meaning of the layer names. By understanding the chemical text rather than just memorizing strings, SciBERT+attention outperforms the best tree ensemble (CatBoost with native categorical handling) by 0.022% ± 0.057% and the gradient-boosted baseline by 0.105% ± 0.071% (Table S7). Tree ensembles fall behind at this level because they cannot cross-reference the naming conventions and material choices of different laboratories, whereas reading what the names actually mean enables true extrapolation.

This regime shift is most pronounced when predicting future discoveries (the temporal split), where the test set is largely unprecedented. In these later years, researchers introduce novel transport layers, distinct compositions, and new deposition procedures from institutions absent in the training data (Table S8). These future devices are consequently more efficient, with a mean PCE of 13.59% against 10.54% in training. On this split, the two architecture families completely diverge (details in temporal split of Methods, Table S7): every neural model improves on the gradient-boosted baseline, by 0.19% to 0.27%, and is more accurate than every tree-based model. We evaluated two temporal splits, one training on publications up to 2016 and testing on those from 2018 onward, the other training on publications up to 2017 and testing on those from 2019 onward, and both place the neural models ahead of the tree ensembles (Table S9). This massive advantage stems directly from the bounds of the models' predictions. Tree ensembles systematically under-predict the test devices, particularly those with novel compositions; none predict above 19.4%, as tree predictions are fundamentally constrained to averaging training-era outcomes. In contrast, neural models are unbounded and successfully follow the physical rise of the field, exhibiting a minimal average bias and predicting efficiencies up to 21.8%. Removing each model's mean bias—a single constant shift per model—brings the two families level across the test set. Therefore, neural architectures successfully extrapolate the field's rising efficiency onto novel devices and materials introduced after the training years, whereas tree ensembles inherently cannot. Ultimately, benchmarking on random splits actively misleads the field by rewarding models that merely memorize the past, while masking their systematic collapse when faced with new institutions and future discoveries.

**Feature importance: from labels to physical properties**

To uncover the physical mechanisms behind this regime shift, we measured which inputs carry the predictive information and how this changes across the evaluation hierarchy (Fig. 4a,b; details in feature importance of Methods). Removing one group of inputs at a time and retraining the gradient-boosting reference model shows that, at level 3, the fabrication process, the device stack, and the absorber composition are all highly predictive, raising the test MAE by 0.091%, 0.082%, and 0.074% respectively when removed (Table S10). Therefore, rather than focusing solely on the absorber chemistry, true physical discovery requires treating the composition, the device architecture, and the fabrication methodology as equally critical pillars that must be co-engineered.

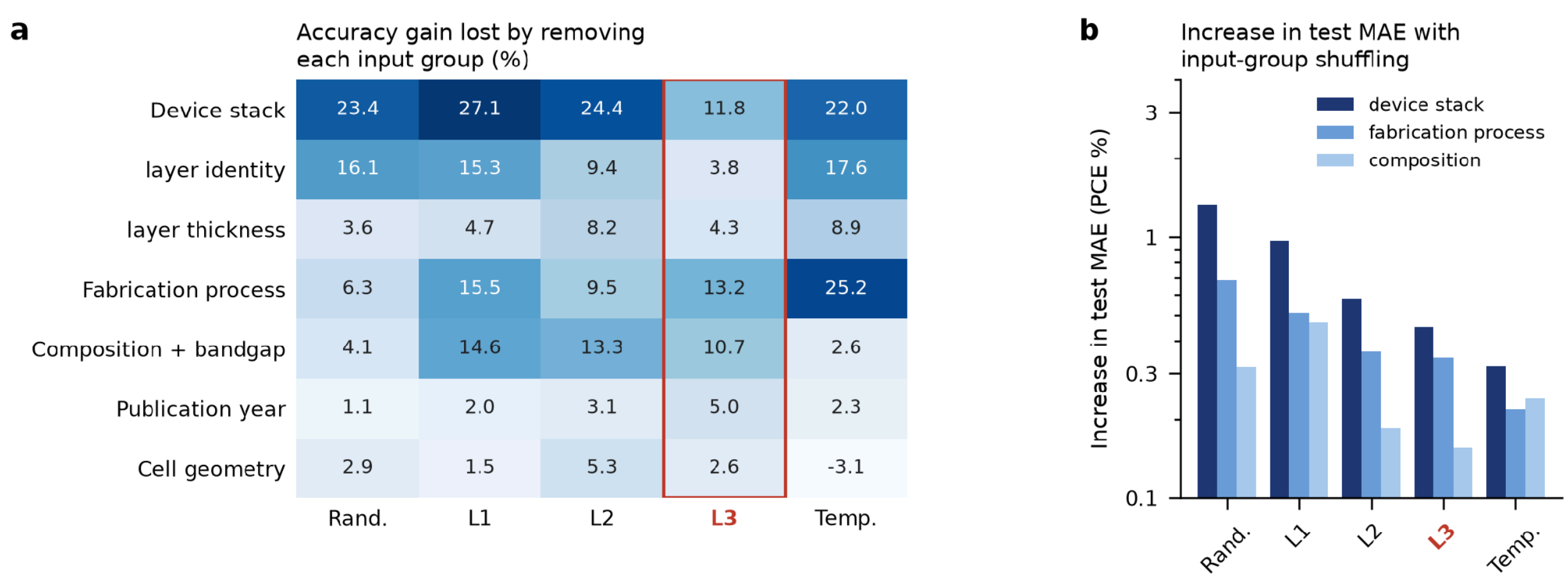

**Origin of the predictive information. a**, The share of the accuracy gain (the improvement of the full model over predicting the training median) that is lost when one group of inputs is removed from a gradient-boosting model, on every split. The device stack combines layer identity and layer thickness; the increases in test MAE are in Table S10. **b**, Increase in the test MAE of the gradient-boosted baseline when the device stack, the fabrication process or the composition is shuffled among the test devices, one group at a time (log scale). Data are in Tables S10 and S11.

To find which inputs lose their value as recall is prevented, we compared the removals across the splits (Fig. 4a; Table S10). On the random split, removing the device stack causes a massive error increase of 0.342%, far exceeding the contribution of either the fabrication process (0.092%) or the composition (0.060%). Most of this inflated value comes specifically from layer identity: removing it raises the error by 0.236% on the random split but only by 0.026% at level 3, a ninefold collapse. In contrast, the fabrication process and the composition keep a contribution of at least 0.06% across the random split and levels 1 to 3 (0.08% to 0.19% and 0.06% to 0.18%). Separating publications and institutions therefore removes most of the value of the recorded layer names as categories, because a layer's proprietary name acts as a categorical barcode that identifies a laboratory and recalls its efficiencies. The fabrication process and the composition, however, keep their genuine physical value. This explains why the model that reads what the layer names mean, rather than treating them as categorical labels, leads at level 3 (Section 3.2).

Shuffling inputs among the test devices with the trained baseline held fixed reveals how the model relies on different features (Table S11). At level 3, shuffling the device stack and the fabrication process raises the error by 0.454% and 0.345%, respectively. As leakage is removed across the grouped splits, the stack loses its predictive value fastest: its error penalty collapses by 65.8% from the random split to level 3 against 49.8% for the fabrication process and 51.1% for the composition. Consequently, the stack's lead over the fabrication process narrows from 1.94 to 1.32 times, proving that its dominance on looser splits is heavily driven by recall, with layer names functioning as categorical laboratory identifiers. On later publications, the model's reliance flattens: shuffling the stack, the composition, and the fabrication process raises the error by 0.320%, 0.241%, and 0.219%. This is consistent with half of the later devices carrying an input absent from training (Table S8), creating a level offset in the baseline's predictions that no shuffling can change. Set against feature-group removal (Table S10), reliance and true physical information completely part ways on this temporal split: the fabrication process, on which the trained model leans least, is the exact group whose removal costs the most information (+0.284%, or 25.2% of the accuracy gain). While these empirical data suggest that the device stack and fabrication methodology impact efficiency more than the absorber composition (which costs only +0.029% or 2.6% when removed), this is a structural limitation of the baseline model. Tree-based models lack a fundamental physical understanding of chemistry, encoding complex material compositions merely as statistical variables. Therefore, while how a device is fabricated undeniably carries critical, irreplaceable information for predicting future discoveries, further study using composition-aware neural architectures is required to successfully model the underlying physics of the absorber.

## Conclusion

Random splits actively misdirect the field of materials science by mistaking memorization for physical understanding. By auditing the evaluation of machine learning models on the Perovskite Database, we demonstrated that standard random splits systematically leak data through massive internal duplication, artificially inflating model performance through the mere recall of historical recipes.

To correct this, we established a rigorous, provenance-based evaluation framework that progressively closes these leakage pathways. By isolating training data at the publication and institution levels, and by testing strictly forward in time, we expose the true extrapolative limits of machine learning models. With this framework in place, we revealed a critical regime shift: while models may perform similarly on weak splits, neural architectures significantly outperform tree-based models in true extrapolation to unseen institutions and future discoveries. We release this entire framework to the community—including the curated data, the mapped institutional provenance, the standardized grouped splits, and the fully benchmarked suite of 13 models.

Moving forward, we recommend adopting four fundamental practices. First, claims of generalisation or physical discovery are best supported by evaluation on level 3 or the temporal split. Second, tree ensembles can serve as standard, rigorous baselines against which new architectures are evaluated. Third, comparisons should be controlled against naive lookup-table baselines to quantify any remaining memorisation. Finally, having shown that the device architecture and fabrication methodology dictate final efficiency as strongly as the absorber chemistry, future machine-learning frameworks should aim to move beyond isolated material predictions and co-engineer all three pillars of the physical device.

**Acknowledgements.**
The authors used the large language models Claude (Anthropic) and Gemini (Google) as assistants in drafting parts of the manuscript and the analysis code. All text, code and results were reviewed, verified and revised by the authors, who take full responsibility for the content of this article.

**Competing interests.**
The authors declare no competing interests.

## Supporting Information

The nineteen core fields and the percentage of the 42,443 original archived records that are missing a value for each field.

| Field | Missing |
|---|---|
| PCE | 2.2% |
| $V_{oc}$ | 5.0% |
| FF | 5.4% |
| $J_{sc}$ | 4.9% |
| Bandgap | 24.9% |
| ETL stack | 0.8% |
| HTL stack | 6.3% |
| Back-contact stack | 0.1% |
| Substrate stack | <0.1% |
| *A*-site ions | 0.1% |
| *A*-site coefficients | 0.1% |
| *B*-site ions | <0.1% |
| *B*-site coefficients | 0.1% |
| *X*-site ions | <0.1% |
| *X*-site coefficients | 0.1% |
| Cell architecture | <0.1% |
| Measured cell area | 1.7% |
| Anneal temperature | 8.1% |
| Deposition procedure | 0.3% |

The 24 descriptive fields of each device, with their source (taken from the archive or derived) and the archive field(s) from which each is built.

| Field | Source | Archive field(s) and construction |
|---|---|---|
| *Absorber* | | |
| Composition as weights of pure compounds | derived | `Perovskite_composition_{a,b,c}_ions` and `_coefficients`: weights of the pure $ABX_3$ compounds in the mixture |
| Compounds outside the compound library | derived | the same composition fields |
| Cations outside the 3D perovskite | derived | `Perovskite_composition_a_ions` and `_coefficients`: *A*-site ions that do not form a 3D perovskite |
| Bandgap | archive | `Perovskite_band_gap` |
| *Device* | | |
| Electron transport layer (text) | derived | `ETL_stack_sequence` and `ETL_additives_compounds`, joined as text |
| Hole transport layer (text) | derived | `HTL_stack_sequence` and `HTL_additives_compounds`, joined as text |

| Field | Source | Archive field(s) and construction |
|---|---|---|
| Back contact (text) | derived | Backcontact_stack_sequence and Backcontact_additives_compounds, joined as text |
| Substrate (text) | derived | Substrate_stack_sequence |
| Architecture | archive | Cell_architecture |
| Flexible | archive | Cell_flexible |
| Semitransparent | archive | Cell_semitransparent |
| Perovskite thickness | archive | Perovskite_thickness |
| Electron transport layer thickness | archive | ETL_thickness |
| Hole transport layer thickness | archive | HTL_thickness_list |
| Back-contact thickness | archive | Backcontact_thickness_list |
| *Fabrication* | | |
| Deposition (text) | derived | Perovskite_deposition_procedure and Perovskite_deposition_solvents, joined as text |
| Anneal temperature | derived | Perovskite_deposition_thermal_annealing_temperature: its first number |
| Anneal time | archive | Perovskite_deposition_thermal_annealing_time |
| Antisolvent quenching | derived | Perovskite_deposition_quenching_induced_crystallisation, as 0 or 1 |
| Solvent annealing | archive | Perovskite_deposition_solvent_annealing |
| *Measurement and record* | | |
| Cell area | archive | Cell_area_measured |
| Cell area, logarithm | derived | the same field: $log(1 + area)$ of its first value |
| Scan direction | archive | JV_default_PCE_scan_direction |
| Publication date | archive | Ref_publication_date |
| *Measured outputs (not part of the description)* | | |
| Power conversion efficiency | archive | JV_default_PCE |
| Open-circuit voltage | archive | JV_default_Voc |
| Short-circuit current density | archive | JV_default_Jsc |
| Fill factor | archive | JV_default_FF |

Unit kept together by each split and the share of test devices with a duplicate, a device from the same publication or a device from the same institution in training (institution share among devices with a known institution), for the published partition of each split.

| Split | Unit kept together | Duplicate | Same publication | Same institution |
|---|---|---|---|---|
| Random | none | 56.9% | 95.8% | 99.8% |
| Level 1 | duplicate group | 0 | 74.9% | 98.5% |
| Level 2 | publication | 0.0% | 0 | 93.0% |
| Level 3 | institution | 0.0% | 0 | 0 |
| Temporal | publication, by date | 0[a] | 0 | 74.1% |

skip [a] The publication date is one of the inputs, so duplicates cannot cross a date boundary.

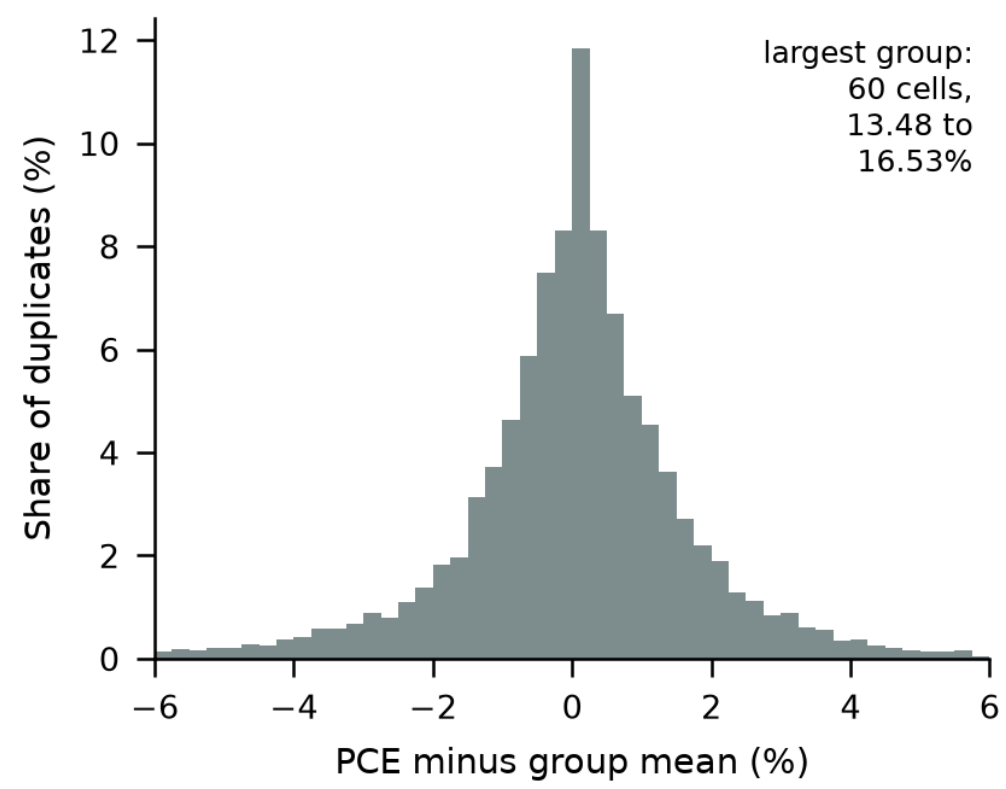

Spread of efficiency within duplicate groups. The PCE of each duplicated device minus the mean PCE of its duplicate group. In the largest group, the PCE of 60 cells ranges from 13.48% to 16.53%.

Test MAE (%) on the replicated evaluation setup of the published state of the art [16] over ten random splits, given as the mean and the standard deviation (SD) over splits.

| Model | Mean MAE ± SE | SD |
|---|---|---|
| CatBoost, default | 2.893 ± 0.011 | 0.035 |
| LightGBM, default | 2.894 ± 0.010 | 0.032 |
| Histogram gradient boosting, default | 2.899 ± 0.011 | 0.035 |
| XGBoost, default | 2.918 ± 0.010 | 0.032 |
| Random forest, default | 2.949 ± 0.009 | 0.029 |
| LightGBM, selected | 2.872 ± 0.011 | 0.034 |
| XGBoost, selected | 2.876 ± 0.012 | 0.037 |
| CatBoost, selected | 2.888 ± 0.012 | 0.038 |
| Lookup table | 2.934 ± 0.015 | 0.048 |
| Reported state of the art [16] | 2.936 | |

Test MAE (%) of the recall controls on every split, for predicting the training median, the lookup table, the nearest-neighbour model (with its *k*) and the gradient-boosted baseline, and each control minus the baseline.

| Split | Training median | Lookup table | Nearest neighbours (*k*) | Baseline | Lookup − baseline | Neighbours − baseline |
|---|---|---|---|---|---|---|
| Random | 3.771 | 2.643 | 2.396 (3) | 2.131 | +0.512 ± 0.049 | +0.264 ± 0.043 |
| Level 1 | 3.793 | 3.793 | 2.850 (20) | 2.457 | +1.336 ± 0.137 | +0.394 ± 0.050 |

| Split | Training median | Lookup table | Nearest neighbours ($k$) | Baseline | Lookup − baseline | Neighbours − baseline |
|---|---|---|---|---|---|---|
| Level 2 | 3.697 | 3.696 | 3.084 (100) | 2.869 | +0.827 ± 0.070 | +0.215 ± 0.055 |
| Level 3 | 3.715 | 3.714 | 3.161 (100) | 3.017 | +0.698 ± 0.091 | +0.144 ± 0.043 |
| Temporal | 4.560 | 4.560 | 3.635 (50) | 3.462 | +1.098 ± 0.055 | +0.173 ± 0.035 |

Range over the ten further partitions of the random split and of each level of the number of test devices and test units (duplicate groups, publications or institutions), and of the share of test devices with a duplicate, a device from the same publication or a device from the same institution in training.

| Split | Test devices | Test units | Duplicate | Same publication | Same institution |
|---|---|---|---|---|---|
| Random | 2,362–2,362 | 2,051–2,109 | 58.0–61.0% | 95.6–96.5% | 99.3–99.8% |
| Level 1 | 2,365–2,387 | 1,260–1,340 | 0.0–0.0% | 72.0–79.9% | 97.2–98.9% |
| Level 2 | 4,724–4,732 | 917–978 | 0.1–0.3% | 0.0–0.0% | 90.2–92.0% |
| Level 3 | 4,528–4,680 | 109–191 | 0.0–0.0% | 0.0–0.0% | 0.0–0.0% |

Paired differences in test MAE (%) between models. **a**, Each comparison model minus the gradient-boosted baseline on every split (negative means the comparison model is more accurate). **b**, Each model minus the lowest-error model on the random split and at each level (positive means less accurate; a dash marks the lowest-error model).

| Model | Random | Level 1 | Level 2 | Level 3 | Temporal |
|---|---|---|---|---|---|
| **a**, minus the gradient-boosted baseline | | | | | |
| XGBoost | −0.104 ± 0.013 | −0.061 ± 0.016 | −0.037 ± 0.018 | −0.067 ± 0.018 | +0.009 ± 0.017 |
| LightGBM | −0.158 ± 0.018 | −0.048 ± 0.026 | −0.021 ± 0.018 | −0.060 ± 0.019 | +0.014 ± 0.017 |
| CatBoost (encoded) | −0.004 ± 0.014 | +0.024 ± 0.020 | +0.022 ± 0.028 | +0.002 ± 0.028 | −0.074 ± 0.024 |
| CatBoost (native) | +0.039 ± 0.018 | +0.080 ± 0.027 | −0.030 ± 0.032 | −0.083 ± 0.036 | −0.022 ± 0.026 |
| Histogram gradient boosting | −0.106 ± 0.013 | −0.028 ± 0.020 | −0.006 ± 0.017 | −0.016 ± 0.015 | +0.060 ± 0.013 |
| Random forest | −0.088 ± 0.021 | +0.024 ± 0.036 | +0.054 ± 0.028 | +0.005 ± 0.028 | +0.159 ± 0.026 |
| MLP | −0.067 ± 0.027 | +0.226 ± 0.085 | +0.068 ± 0.044 | −0.024 ± 0.046 | −0.269 ± 0.043 |
| ResNet-MLP | −0.047 ± 0.029 | +0.261 ± 0.065 | +0.164 ± 0.045 | +0.015 ± 0.049 | −0.193 ± 0.039 |
| FT-Transformer | −0.053 ± 0.026 | +0.244 ± 0.071 | +0.092 ± 0.055 | −0.040 ± 0.049 | −0.247 ± 0.038 |
| TabM | −0.091 ± 0.023 | +0.088 ± 0.066 | +0.095 ± 0.037 | −0.038 ± 0.041 | −0.216 ± 0.034 |
| SciBERT+MLP | −0.051 ± 0.029 | +0.107 ± 0.051 | +0.032 ± 0.061 | −0.052 ± 0.072 | −0.202 ± 0.044 |
| SciBERT+attention | −0.139 ± 0.029 | −0.003 ± 0.046 | +0.002 ± 0.054 | −0.105 ± 0.071 | −0.233 ± 0.048 |
| **b**, minus the lowest-error model | | | | | |
| Gradient-boosted baseline | +0.158 ± 0.018 | +0.061 ± 0.016 | +0.037 ± 0.018 | +0.105 ± 0.071 | |
| XGBoost | +0.054 ± 0.012 | — | — | +0.038 ± 0.063 | |

| Model | Random | Level 1 | Level 2 | Level 3 | Temporal |
|---|---|---|---|---|---|
| LightGBM | — | +0.013 ± 0.018 | +0.016 ± 0.016 | +0.045 ± 0.064 | |
| CatBoost (encoded) | +0.154 ± 0.019 | +0.085 ± 0.021 | +0.060 ± 0.021 | +0.107 ± 0.066 | |
| CatBoost (native) | +0.198 ± 0.023 | +0.141 ± 0.027 | +0.007 ± 0.026 | +0.022 ± 0.057 | |
| Histogram gradient boosting | +0.052 ± 0.011 | +0.033 ± 0.018 | +0.032 ± 0.020 | +0.089 ± 0.075 | |
| Random forest | +0.070 ± 0.016 | +0.085 ± 0.032 | +0.092 ± 0.023 | +0.110 ± 0.078 | |
| MLP | +0.092 ± 0.024 | +0.287 ± 0.082 | +0.106 ± 0.038 | +0.081 ± 0.063 | |
| ResNet-MLP | +0.111 ± 0.025 | +0.322 ± 0.062 | +0.201 ± 0.041 | +0.120 ± 0.063 | |
| FT-Transformer | +0.105 ± 0.021 | +0.305 ± 0.068 | +0.130 ± 0.051 | +0.065 ± 0.053 | |
| TabM | +0.068 ± 0.020 | +0.149 ± 0.064 | +0.133 ± 0.035 | +0.067 ± 0.053 | |
| SciBERT+MLP | +0.107 ± 0.026 | +0.168 ± 0.049 | +0.069 ± 0.054 | +0.053 ± 0.048 | |
| SciBERT+attention | +0.019 ± 0.024 | +0.058 ± 0.045 | +0.040 ± 0.047 | — | |

Share of the test devices of the two temporal splits (Table S9) whose value of each field never occurs in the training devices, or in the training and validation devices, of the same split. Institution shares are of the test devices with a known institution.

| Field | Absent from training | Absent from training and validation |
|---|---|---|
| *Split 1: training to 2016, validation 2017, test from 2018 (10,752 test devices)* | | |
| Electron-transport layer | 21.1% (2,265) | 19.0% (2,039) |
| Hole-transport layer | 18.5% (1,984) | 17.7% (1,898) |
| Deposition procedure | 6.7% (716) | 5.5% (593) |
| Absorber composition | 9.5% (1,017) | 7.9% (853) |
| Additive | 2.3% (246) | 1.9% (201) |
| Back contact | 2.2% (237) | 1.8% (194) |
| Substrate | 1.2% (131) | 1.1% (119) |
| Architecture | 0.1% (9) | 0.0% (0) |
| At least one of the above | 49.7% (5,340) | 45.2% (4,863) |
| Full device description | 100.0% (10,752) | 100.0% (10,752) |
| Corresponding institution | 25.9% (2,706) | 16.4% (1,714) |
| *Split 2: training to 2017, validation 2018, test from 2019 (5,079 test devices)* | | |
| Electron-transport layer | 21.3% (1,084) | 19.0% (964) |
| Hole-transport layer | 21.6% (1,097) | 18.7% (950) |
| Deposition procedure | 5.3% (269) | 3.3% (169) |
| Absorber composition | 9.3% (473) | 7.6% (386) |
| Additive | 2.3% (116) | 2.3% (116) |
| Back contact | 1.4% (70) | 1.4% (70) |
| Substrate | 1.3% (65) | 1.3% (65) |
| Architecture | 0.0% (0) | 0.0% (0) |
| At least one of the above | 49.5% (2,516) | 44.1% (2,238) |
| Full device description | 100.0% (5,079) | 100.0% (5,078) |
| Corresponding institution | 19.3% (963) | 13.2% (660) |

Test MAE (%) of each model on the two temporal splits, trained on publications up to 2016 and tested on those from 2018 onward (the temporal split of Table 1 in the main text), and trained on publications up to 2017 and tested on those from 2019 onward.

| Model | Type | To 2016 / from 2018 | To 2017 / from 2019 |
|---|---|---|---|
| MLP | neural | 3.192 | 3.040 |
| FT-Transformer | neural | 3.215 | 3.040 |
| SciBERT+attention | neural | 3.228 | 3.019 |
| TabM | neural | 3.246 | 3.106 |
| SciBERT+MLP | neural | 3.259 | 3.044 |
| ResNet-MLP | neural | 3.268 | 3.060 |
| CatBoost (encoded) | tree | 3.388 | 3.222 |
| CatBoost (native) | tree | 3.440 | 3.226 |
| Gradient-boosted baseline | baseline | 3.462 | 3.175 |
| XGBoost | tree | 3.471 | 3.271 |
| LightGBM | tree | 3.476 | 3.241 |
| Histogram gradient boosting | tree | 3.521 | 3.268 |
| Random forest | tree | 3.621 | 3.500 |

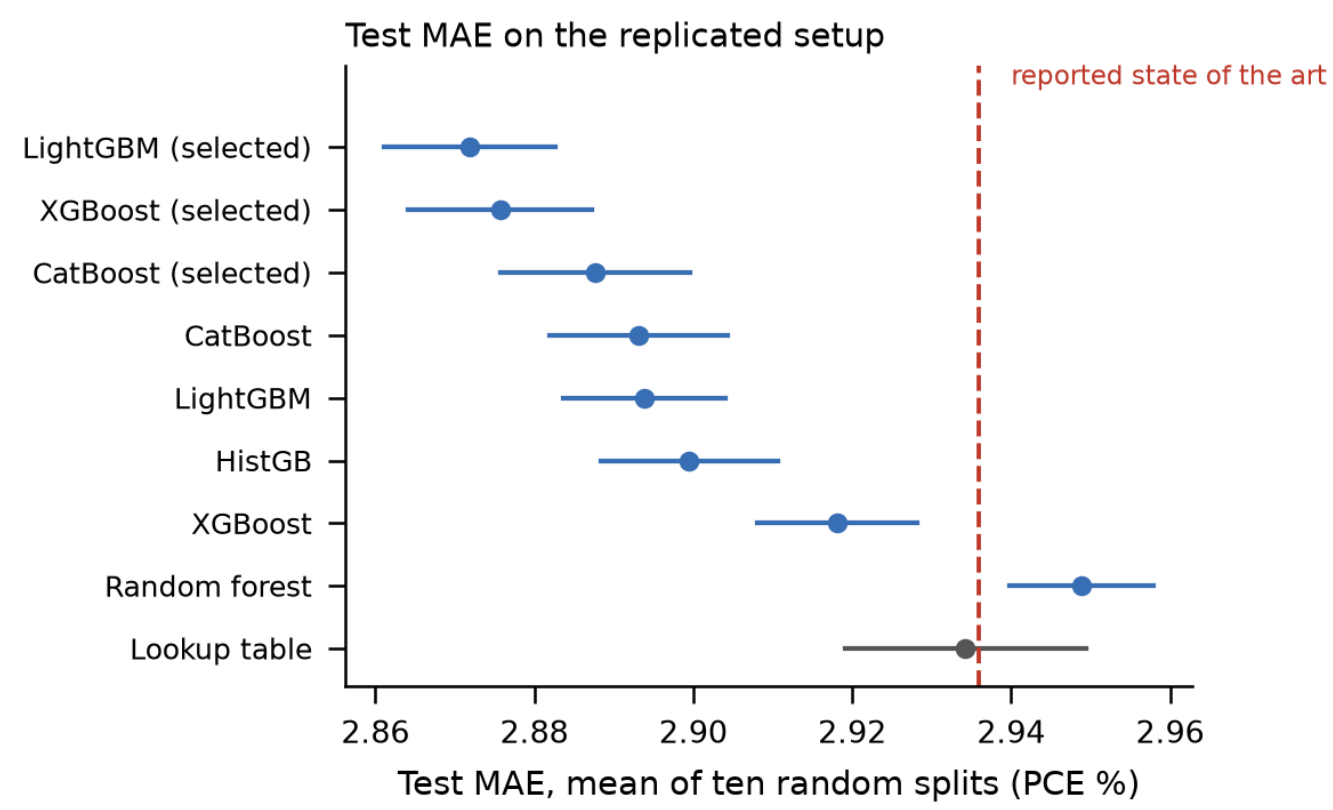

Tree ensembles evaluated on the published state-of-the-art setup [16] (28,922 devices, the 34 most frequent absorber formulas, the names of the five device layers as the only inputs). Mean test MAE over ten random 80/10/10 splits, ± the standard error of the mean, for tree ensembles with their default settings, three with settings chosen on the validation set (selected), and the lookup table. The dashed line marks the reported state of the art (2.936%) [16].

Feature-group removal on every split. **a**, Increase in test MAE (%) when one group of inputs is removed from the gradient-boosting model; the device stack combines layer identity and layer thickness. **b**, The same increase as a share of the accuracy gain (the improvement of the full model over predicting the training median).

| Group removed | Random | Level 1 | Level 2 | Level 3 | Temporal |
|---|---|---|---|---|---|
| **a**, increase in test MAE (%) | | | | | |
| Device stack | 0.342 ± 0.032 | 0.337 ± 0.077 | 0.201 ± 0.059 | 0.082 ± 0.057 | 0.249 ± 0.043 |
| layer identity | 0.236 ± 0.028 | 0.190 ± 0.067 | 0.077 ± 0.060 | 0.026 ± 0.046 | 0.198 ± 0.036 |
| layer thickness | 0.053 ± 0.013 | 0.059 ± 0.027 | 0.067 ± 0.023 | 0.030 ± 0.022 | 0.100 ± 0.019 |
| Fabrication process | 0.092 ± 0.022 | 0.193 ± 0.051 | 0.078 ± 0.034 | 0.091 ± 0.046 | 0.284 ± 0.041 |
| Composition and bandgap | 0.060 ± 0.017 | 0.181 ± 0.102 | 0.110 ± 0.026 | 0.074 ± 0.029 | 0.029 ± 0.029 |
| Publication year | 0.017 ± 0.011 | 0.025 ± 0.023 | 0.026 ± 0.014 | 0.035 ± 0.020 | 0.026 ± 0.012 |

| Group removed | Random | Level 1 | Level 2 | Level 3 | Temporal |
|---|---|---|---|---|---|
| Cell geometry | 0.043 ± 0.011 | 0.019 ± 0.017 | 0.044 ± 0.017 | 0.018 ± 0.016 | −0.034 ± 0.017 |
| Accuracy gain | 1.465 | 1.245 | 0.824 | 0.694 | 1.128 |
| **b**, share of the accuracy gain | | | | | |
| Device stack | 23.4% | 27.1% | 24.4% | 11.8% | 22.0% |
| layer identity | 16.1% | 15.3% | 9.4% | 3.8% | 17.6% |
| layer thickness | 3.6% | 4.7% | 8.2% | 4.3% | 8.9% |
| Fabrication process | 6.3% | 15.5% | 9.5% | 13.2% | 25.2% |
| Composition and bandgap | 4.1% | 14.6% | 13.3% | 10.7% | 2.6% |
| Publication year | 1.1% | 2.0% | 3.1% | 5.0% | 2.3% |
| Cell geometry | 2.9% | 1.5% | 5.3% | 2.6% | -3.1% |

Feature reliance via permutation importance on every split. Increase in the test MAE (%) of the gradient-boosted baseline when the composition inputs, the device-stack inputs or the fabrication-process inputs are shuffled among the test devices, one group at a time. The values average over the model's seeds (five on the random split, three on the others) with five shuffles per seed. The groups correspond to those in Table S10; publication year and cell geometry are not shuffled.

| Inputs shuffled | Random | Level 1 | Level 2 | Level 3 | Temporal |
|---|---|---|---|---|---|
| Composition and bandgap | 0.318 | 0.471 | 0.186 | 0.156 | 0.241 |
| Device stack: layer identity and thickness | 1.329 | 0.969 | 0.580 | 0.454 | 0.320 |
| Fabrication process | 0.687 | 0.512 | 0.364 | 0.345 | 0.219 |

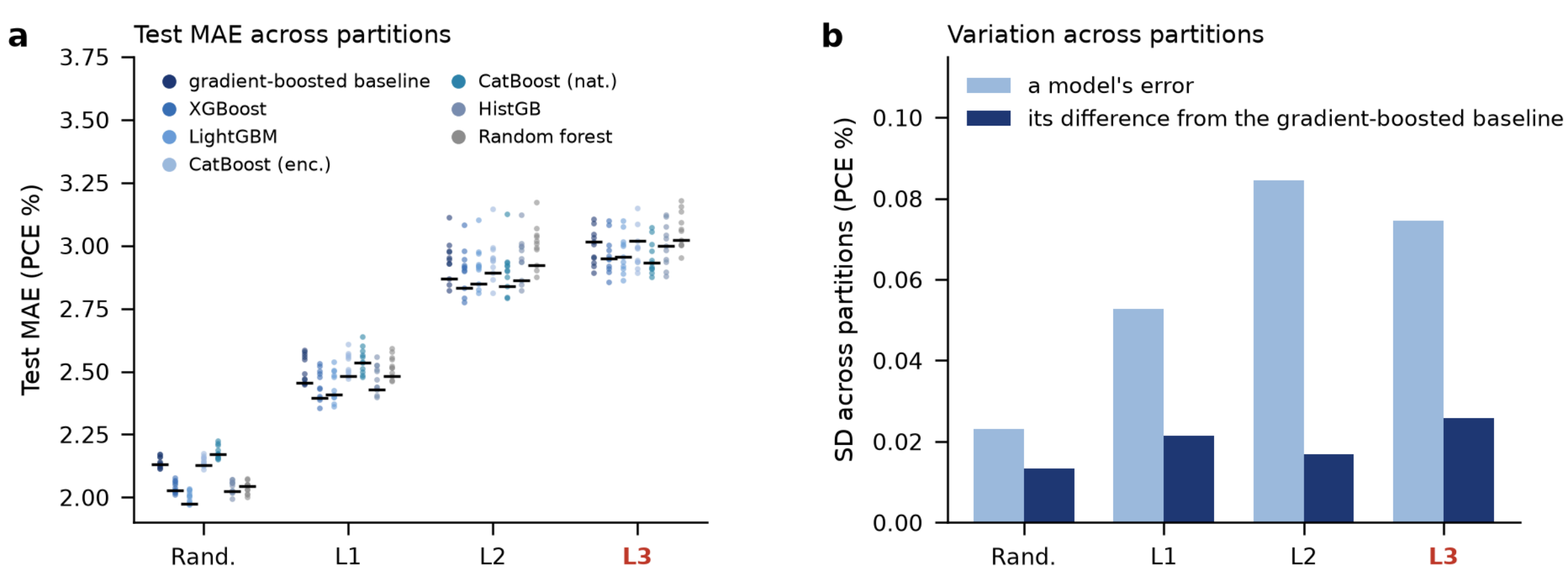

The six tree-based models and the baseline on eleven partitions of each split. **a**, Test MAE on the published partition (black bars) and on ten further partitions drawn with the same rule. **b**, Standard deviation across the eleven partitions of a model's test MAE, and of its difference from the gradient-boosted baseline on the same partition, averaged over the tree-based models.

## Methods

### Data curation.

Device records are from the Perovskite Database [13], which distributes 42,443 records, of which 23,620 survive three filtering stages. First, a record must carry all 19 core device fields (Table S1; the four J-V outputs, the reported gap, the four stack sequences, the six composition fields, architecture, measured area, anneal temperature and deposition procedure). Second, records are removed when a measured output is implausibly low (PCE ≤ 0.1%, $V_{oc} \leq 0.2$ V, FF ≤ 0.25 or $J_{sc} \leq 1$ mA cm$^{-2}$), when the record describes a tandem device or a module, or when its composition cannot be parsed into fixed ion proportions or places more than 15% of its weight on pure $ABX_3$ compounds that cannot be resolved. Third, records are dropped when a label is physically unusable (FF outside (0,1], $J_{sc}$ more than 1.5 mA cm$^{-2}$ above the Shockley–Queisser limit at the recorded bandgap, $V_{oc}$ above the bandgap minus 0.25 V, light intensity outside 95–105 mW cm$^{-2}$, PCE inconsistent with $V_{oc}J_{sc}$FF by more than 2%) or when the record is a mini-module. Each surviving record is expanded into the 24 descriptive fields of Table S2.

### Model inputs.

The tree-based models receive 40 columns built from 23 of the 24 fields, all but the compounds outside the compound library, which the text models read. Each text field, such as the recorded name of a transport layer, is treated as a category: the whole text is one label, and its words are not read. For each such field the models receive the mean efficiency of the training devices with the same text, shrunk towards the overall training mean with a smoothing weight of 20 training devices (target encoding, also called mean encoding [25, 26]), and the number of training devices with that text (count encoding). The encoding of a training device is computed out of fold, from the other four of five folds of the training set [26]; validation and test devices are encoded from the whole training set, so no test label enters the encoding. A text that never occurs in training receives the training mean and a count of zero. The other categorical fields, such as the architecture, are encoded in the same way. Numerical fields are given directly; the seven parsed from the archive's free text (the four layer thicknesses, the anneal time, the measured cell area and the publication date) each come with a flag marking a missing value, and missing values are left missing rather than imputed. The four tabular neural models receive the same 40 columns together with the weights of the absorber over pure compounds, with missing values set to the training median and a quantile transformation fitted on the training devices. The two text models read the text fields through the frozen SciBERT encoder [24], applied once to each text and cached, together with the numerical columns: SciBERT+MLP receives the mean-pooled embedding of each field, and SciBERT+attention pools its token states with learned attention. Because the encoder is frozen, nothing is fitted to the corpus.

**Provenance mapping.**

The publication of each device is identified by the DOI in its original record, normalised to lower case without surrounding spaces. The archive records no laboratory identifier; its only related field is the surname of the lead author, and "Wang et al." alone covers 1,199 devices from unrelated groups. The author affiliations of every publication were therefore retrieved from OpenAlex by DOI, in batches of 50, and the institution of each publication is that of its corresponding author. OpenAlex resolves 99.9% of the 4,683 publications and gives the corresponding-author institution of 94.7% of them, 796 institutions covering 95.8% of the devices. A publication that OpenAlex does not resolve, or that has no corresponding author, is recorded without an institution and is never assigned one by name.

**Duplication analysis.**

Two devices are duplicates when they agree in all 24 descriptive fields. Duplicates form groups, and each group is traced back to the original archive records to test whether its devices are identical in all 249 archive columns that describe how a device was made (every column except measured outcomes and bibliographic fields). The spread of efficiency within a group is measured as the PCE of each duplicated device minus the mean PCE of its group (Fig. S1).

**Split definitions.**

Every model is trained and scored on the same five splits, each of which assigns whole units to training, validation and test (Table S3). The random split assigns single devices in the proportions 80/10/10, stratified by PCE quintile (18,896/2,362/2,362 training, validation and test devices). Level 1, feature-grouped, assigns each group of duplicates whole, stratified by group-mean PCE and packed by cumulative device count to 80/10/10 (18,882/2,363/2,375 devices; 1,258 test groups). At level 2, publication-grouped, and level 3, institution-grouped, publications or corresponding-author institutions are shuffled and assigned whole, to the test set until it holds 20% of the devices, then to validation until it holds 10%, and the rest to training; the larger test fraction holds out enough publications or institutions for a stable evaluation. Level 2 has 16,492/2,370/4,758 devices from 3,327/491/878 publications; the 25 test devices of five publications whose DOI also occurs in training in another spelling are not scored. Level 3 has 15,535/2,275/4,812 devices from 542/85/169 institutions; the 998 devices without a known institution are left out rather than added to training. The temporal split assigns each publication by its publication date: publications to 2016 to training, 2017 to validation and from 2018 to test (8,076/4,792/10,752 devices; 1,642 training and 2,071 test publications). A publication carries one date, so none crosses the boundary. The published partition of each split uses seed 42. Ten further partitions of the random split and of each level use seeds 1 to 10 with the same rule; rebuilding seed 42 with the rule reproduces each published partition exactly. Target normalisation constants are computed from training devices only.

**Recall controls.**

Two controls predict by recalling training devices rather than by learning how efficiency depends on the inputs. The lookup table predicts, for each test device, the median PCE of the training devices with identical inputs, and the training median when there is none; the median is the constant prediction with the lowest absolute error. The nearest-neighbour model predicts the mean PCE of the $k$ training devices closest to the test device in the baseline's inputs, standardised on the training devices with missing values set to the training median, with $k$ chosen on validation MAE from 1, 3, 5, 10, 20, 50, 100, 200 and 500.

**Models.**

The gradient-boosted baseline is HistGradientBoostingRegressor with absolute-error loss, learning rate 0.03, 1500 maximum iterations, 63 leaves and early stopping on a validation fraction of the training devices, averaged over three seeds. The twelve comparison models are XGBoost, LightGBM, CatBoost [27] with out-of-fold target encoding and with its native categorical handling, which uses its own form of target encoding (ordered target statistics), a histogram gradient-boosting model, a random forest, an MLP, a ResNet-style MLP, an FT-Transformer, TabM, SciBERT+MLP and SciBERT+attention. A boosted reference model, lighter than the baseline (400 iterations, learning rate 0.06, three seeds), is used where a model must be refitted many times: in the leakage analysis, the training-size controls and feature-group removal.

**Benchmark settings.**

The configuration of the baseline was selected once, against level-1 validation error. Each comparison model was optimised once with a tree-structured Parzen estimator on level 1 alone, minimising validation MAE. Each selected configuration is applied unchanged at every split, with three seeds averaged; no test device is used to tune a configuration, to stop training or to select a model, and every model is scored on the same test devices. On the ten further partitions, the six tree-based models and the baseline are refitted with their published configurations and seeds. The earlier evaluation setup of Solar-GECO [16] is replicated from its description, since its code and processed data were not released: devices whose absorber has a single ion on every site and a recorded PCE are kept, the formula notation is normalised, multi-layer absorbers are removed and the 34 most frequent absorber formulas are retained, giving 28,922 devices from 5,565 publications (the original study reports 29,344 devices). The only inputs are the names of the five device layers (absorber, ETL, HTL, back contact and substrate), given to the tree ensembles as out-of-fold target encodings and counts of each field and of the five-field combination, with CatBoost also receiving them as native categories. Default settings are the library defaults with the absolute-error loss (random forest: the library defaults), fitted on the training devices only; selected settings are chosen from a small grid on the validation devices only. Each model is fitted on ten random 80/10/10 splits with seeds 0 to 9.

**Leakage analysis.**

The effect of related devices is measured with the reference model on training sets that differ in one respect only. For publications, the test devices are half of the devices of each publication in the level-2 test set, and the other half are the related devices; all duplicates of the test devices are removed, and both training sets, with and without the related devices, are completed with 8,000 devices from other publications, so that the test devices and the training-set size are fixed. For institutions, the publications of each of the 101 institutions of the level-3 test set with at least two publications are divided at random into two halves; devices of the first half are test devices (2,096 to 2,273 per repetition) and devices of the second half the related devices (2,247 to 2,420), so that no test publication is in either training set, with duplicates removed, devices without a known institution left out and both training sets completed with 8,000 devices from other institutions. Each comparison is repeated twenty times. Two training-size controls measure the effect of the smaller training sets of the stricter levels: the level-1 training set is shrunk to the size of the level-2 one (three splits), and the training sets of three level-2 partitions (the published one and two further draws) are shrunk to the 15,535 devices of level 3 (five random subsets each).

**Temporal split.**

The temporal split is evaluated under two framings. The main framing trains on publications to 2016, validates on 2017 and tests from 2018 (Split definitions). The second trains on publications to 2017, validates on 2018 and tests from 2019 (12,868/5,673/5,079 devices), to test prediction across a two-year gap. A test device of either split is counted as new in a field when its value of that field never occurs in the training devices of that split, or in its training and validation devices; institution shares are of the test devices with a known institution.

**Feature importance.**

Inputs are assigned to groups by column provenance rather than by name, so that layer thickness is separated from layer identity: the device stack (the identity of the electron- and hole-transport layers, back contact and substrate, and the layer thicknesses), the fabrication process (deposition and additive descriptions, annealing temperature and time, anti-solvent and solvent-annealing treatments), composition and bandgap, publication year, and cell geometry (all other inputs, such as the cell area, architecture and flexibility). Feature-group removal retrains the reference model with one group removed and reports the increase in test MAE, also as a share of the accuracy gain, the full model's improvement over predicting the training median. Shuffling keeps the trained baseline fixed and reassigns one group at a time, the composition and bandgap, the device stack or the fabrication process, at random among the test devices while the other inputs stay in place, five times per seed (five seeds on the random split, three on the others); publication year and cell geometry are not shuffled. Shuffling supplies contradictory rather than absent inputs, so its magnitudes are compared between splits, not with those of removal.

**Uncertainty estimation.**

Uncertainties are given as ± one standard error. For a result on a test set, such as the difference between two models or the rise in error when inputs are removed, the standard error is the width of the 95% interval of a paired bootstrap (20,000 resamples, seed 0) divided by 3.92. The bootstrap resamples each split's grouping unit, never single devices, since devices within a unit are not independent: groups of duplicates for the random split and level 1, publications at level 2 and on the temporal split, and institutions at level 3. For a mean over repeated splits or repetitions (the ten random splits of the earlier setup, the twenty repetitions of each related-device comparison and the three partitions of each training-size control), the standard error is that of the mean. Training-seed variation is reported separately and never pooled with test uncertainty. Unless stated otherwise, ± in tables and figures denotes the standard error defined here.

**Data and code availability.**

A reproducibility release accompanies the paper: the row-level assignments of every split and of the ten further partitions, the publication and OpenAlex-resolved institution of every device, the optimisation trace, selected configuration and test predictions of each of the thirteen models on every split, the scored result tables, and all analysis code from dataset construction to interval reproduction. The Perovskite Database source file is not redistributed; the release includes replay scripts that reconstruct the processed data and splits bit-identically from the public source [13], now distributed through NOMAD (https://nomad-lab.eu/prod/v1/gui/search/perovskite-solar-cells-database), with a sha256 of the exact file used. With the split assignments and the provenance of every device, this is sufficient to place a new model on the same protocol without retraining ours. [Repository DOI to be inserted at submission.]